\documentclass[11pt]{article}
\usepackage{amsfonts,amsmath,amsthm,amssymb,graphicx,mathrsfs}

\usepackage[top=3cm,bottom=3cm,left=3cm,right=3cm]{geometry}
\usepackage[colorlinks=true,urlcolor=blue, linkcolor =blue,citecolor=blue]{hyperref}
\usepackage{booktabs}
\usepackage[width=.9\textwidth, labelfont=bf,
textfont=it]{caption}

\theoremstyle{plain}

\newtheorem{remark}{Remark}
\begin{document}

\title{Simulation comparisons between Bayesian and de-biased estimators in low-rank matrix completion%\thanks{Grants or other notes
}


\author{The Tien Mai}

\date{
	\begin{small}
		Department of Mathematical Sciences, 
		\\
		Norwegian University of Science and Technology,
		7034 Trondheim, Norway.
		\\
		Email: the.t.mai@ntnu.no
	\end{small}
}

%\authorrunning{T. Tien Mai} % if too long for running head

%\institute{T.T. Mai \at
%          Department of mathematical sciences, Norwegian University of Science and Technology, Trondheim, Norway \\
%              \email{the.t.mai@ntnu.no}           %  \\
       %    \and
        %   S. Author \at
         %     second address
%}


\maketitle

\begin{abstract}
	In this paper, we study the low-rank matrix completion problem, a class of machine learning problems, that aims at the prediction of missing entries in a partially observed matrix. Such problems appear in several challenging applications such as collaborative filtering, image processing, and genotype imputation. We compare the Bayesian approaches and a recently introduced de-biased estimator which provides a useful way to build confidence intervals of interest.  From a theoretical viewpoint, the de-biased estimator comes with a sharp minimax-optimal rate of estimation error whereas the Bayesian approach reaches this rate with an additional logarithmic factor. Our simulation studies show originally interesting results that the de-biased estimator is just as good as the Bayesian estimators. Moreover, Bayesian approaches are much more stable and can outperform the de-biased estimator in the case of small samples. In addition, we also find that the empirical coverage rate of the confidence intervals obtained by the de-biased estimator for an entry is absolutely lower than of the considered credible interval. These results suggest further theoretical studies on the estimation error and the concentration of Bayesian methods as they are  quite limited up to present. 
Keywords: \textit{Low-rank matrix, matrix completion, Bayesian method, de-biased estimator, uncertainty quantification, confidence interval.}
%\subclass{MSC code1 \and MSC code2 \and more}
\end{abstract}

\section{Introduction}
The goal of low-rank matrix completion is to recover a low-rank matrix from its partially (noisy) observed entries. This problem has recently received an increased attention due to the emergence of several challenging applications, such as recommender systems \cite{xiong2010temporal,adomavicius2011context} (particularly the famous Netflix challenge \cite{bennett2007netflix}), genotype imputation \cite{chi2013genotype,jiang2016sparrec},  image processing \cite{cabral2014matrix,luo2015multiview,he2014image} and quantum state tomography \cite{gross2010quantum,mai2017pseudo,mai2021effici}. Different approaches from frequentist to Bayesian methods have been proposed and studied from theoretical and computational points of views, see for example \cite{candes2010matrix,candes2009exact,candes2010power,koltchinskii2011nuclear,recht2013parallel,alquier2013bayesian,lawrence2009non,lim2007variational,salakhutdinov2008bayesian,zhou2010nonparametric,mai2015,mai2021bayesian,chen2019inference,alquier2020concentration}.

From a frequentist point of view, most of the recent methods are usually based on penalized optimization. A seminal result can be found in \cite{candes2009exact,candes2010power} for exact matrix completion (noiseless case) and further developed in the noisy case in~\cite{candes2010matrix,koltchinskii2011nuclear,negahban2012restricted}.  Some efficient algorithms had also been studied, for example see~\cite{mazumder2010spectral,recht2013parallel,hastie2015matrix}.  More particularly, in the notable work \cite{koltchinskii2011nuclear}, the authors studied nuclear-norm penalized estimators and provided reconstruction error rate for their methods. They also showed that these error rates are minimax-optimal (up to a logarithmic factor). Note that the error rate, i.e. the average quadratic error
on the entries, of a rank-$r$ matrix size $ m \times p $ from $ n $-observations can not be
better than: $ r\max(m,p)/n $~\cite{koltchinskii2011nuclear}.

More recently, in a work by \cite{chen2019inference}, de-biased estimators have been proposed for the problem of noisy low-rank matrix completion.  The estimation accuracy of this estimator is shown to be sharp in the sense that it reaches the minimax-optimal rate without any additional logarithmic factor.  A sharp bound has also been obtained by a different estimator in \cite{klopp2015matrix}. However, uncertainty quantification is not given. More importantly, the confidence intervals on the reconstruction of entries of the underlying matrix are also provided by using the de-biased estimators in the work by \cite{chen2019inference}. It is noted that conducting uncertainty quantification for matrix completion is not straightforward. This is because, in general, the solutions for matrix completion do not admit closed-form and the distributions of the estimates returned by the state-of-the-art algorithms are hard to derive.

On the other hand,  uncertainty quantification can be obtained straightforwardly from a Bayesian perspective. More specifically, the unknown matrix is considered as a random variable with a specific prior distribution and statistical inference can be obtained using the posterior distribution, for example considering credible intervals.  Bayesian methods have been studied for low-rank matrix completion mainly from a computational viewpoint \cite
{lim2007variational,salakhutdinov2008bayesian,zhou2010nonparametric,alquier2014bayesian,alquier2014bayesian,lawrence2009non,cottet20181,babacan2012sparse,yang2018fast}.  Most Bayesian estimators are based on conjugate priors which allow to use Gibbs sampling \cite{alquier2014bayesian,salakhutdinov2008bayesian} or Variational Bayes methods~\cite{lim2007variational}. These algorithms are fast enough to deal with and actually tested on large datasets like Netflix \cite{bennett2007netflix} or MovieLens \cite{harper2015movielens}. However, the theoretical understanding of Bayesian estimators is quite limited, up to our knowledge, \cite{mai2015} and \cite{alquier2020concentration} are the only prominent examples. More specifically,  they showed that a Bayesian estimator with a low-rank factorization prior reaches the minimax-optimal rate up to a logarithmic factor and the paper \cite{alquier2020concentration} further shows that the same estimation error rate can be obtained by using a Variational Bayesian estimator. 

In this paper, to understand the performance of Bayesian approaches when compared to the de-biased estimators, we perform numerical comparisons on the estimation accuracy (the estimation error, the normalized squared error and the prediction error, see Section \ref{sc_numerical}) considering the de-biased estimator in \cite{chen2019inference} and the Bayesian methods \cite{alquier2020concentration} for which the statistical properties have been well studied.  Furthermore,  we examine in detail the behaviour of the confidence intervals obtained by the de-biased estimator and the Bayesian credible intervals. Interestingly, it is noted that recent works \cite{rendle2019difficulty,rendle2020neural} show that Bayesian methods are now the most accurate in practical recommender systems. Although Bayesian methods have become popular in the problem of  matrix completion, its uncertainty quantification (e.g. credible intervals) has received much more limited attention in the literature.  

Results from simulation comparisons release originally interesting messages. More specifically, the de-biased estimator is just as good as the Bayesian estimators when we look at the estimation accuracy, although it is completely successful in improving the estimator being de-biased. On the other hand, the Bayesian approaches are much more stable than the de-biased method and, in addition, they outperform the de-biased estimator especially in the case of small samples. Moreover, we find that the coverage rates of the 95\% confidence intervals obtained using the de-biased estimator are lower than the 89\% equal-tailed credible intervals. These evidences suggest that the Bayesian estimators may actually reach the minimax-optimal rate sharply and the log-term could be due to the technical proofs (the PAC-Bayesian bounds technique). Furthermore, the concentration rate of the corresponding Bayesian posterior discussed in \cite{alquier2020concentration} with a log-term might not be tight.

The rest of the paper is structured as follows.  In Section \ref{sc_model_lmc} we present the low-rank matrix completion problem, then introduce the de-biased estimator and the corresponding confidence interval and provide details on the considered Bayesian estimators.  In Section \ref{sc_numerical}, simulation studies comparing the different methods are presented.  We discuss our results and give some concluding remarks in the final section.

\section{Low-rank matrix completion}
\label{sc_model_lmc}
\subsection{Model}

In this work, we adopt the statistical model commonly studied in the literature for noisy matrix completion \cite{chen2019inference}. Let $ M^* \in  \mathbb{R}^{m\times p} $ be an unknown rank-$r$ matrix of interest. We partially observe some noisy entries of $M^*$ as 
\begin{equation}
\label{main model}
Y_{ij} = M^*_{ij} + \mathcal{E}_{ij},   \quad (i,j) \in \Omega
\end{equation}
where $\Omega \subseteq \lbrace1, \ldots, m \rbrace\times\lbrace1, \ldots,p \rbrace $ is a small subset of indexes and $ \mathcal{E}_{ij} \sim \mathcal{N}(0, \sigma^2) $ are independently generated noise at the location $(i,j) $. The random sampling model is assumed that each index $ (i,j) \in \Omega $ is observed independently with probability $ \kappa $
(i.e., data are missing uniformly at random). Then, the problem of estimating $ M^* $ with $ n = |\Omega| < mp $ is called the (noisy) low-rank matrix completion problem.

Let $  \mathcal{P}_\Omega (\cdot) : \mathbb{R}^{m\times p}  \mapsto  \mathbb{R}^{m\times p}  $ be the orthogonal projection onto the observed entries in the index set $\Omega $ that
\begin{align*}
\mathcal{P}_\Omega (Y)_{ij} = 
\begin{cases}
Y_{ij},  & \text{ if } (i,j) \in \Omega,
\\
0 ,   & \text{ if } (i,j) \notin \Omega 
\end{cases} .
\end{align*}

\paragraph{Notations:} For a matrix $A\in  \mathbb{R}^{m\times p} $, $ \|A\|_F = \sqrt{{\rm trace}(A^\top A)} $ denotes its Frobenius norm and $ \|A\|_* = {\rm trace} (\sqrt{A^\top A}) $ denotes its nuclear norm. $ \left[ a \pm b\right] $ denotes the interval $  \left[ a -b,  a + b \right] $. We use $ I_q $ to denote the identity matrix of dimension $q \times q $. 

\subsection{The de-biased estimator}
Let $\hat{M}$ be either the solution of the following nuclear norm regularization \cite{mazumder2010spectral}
$$
\min_{Z\in  \mathbb{R}^{m\times p}} \frac{1}{2} \| \mathcal{P}_\Omega (Z - Y) \|_F^2 + \lambda \|Z \|_* ,
$$
or of the following factorization minimization \cite{hastie2015matrix}
\begin{align}
\label{estimator_als}
\min_{U\in  \mathbb{R}^{m\times r} , V\in  \mathbb{R}^{p\times r}  } \frac{1}{2} \| \mathcal{P}_\Omega (Y - UV^\top) \|_F^2 + \frac{\lambda}{2} \|U \|^2_F +  \frac{\lambda}{2} \|V \|^2_F,
\end{align}
where $\lambda>0 $ is a tuning parameter. The optimization problem in \eqref{estimator_als} can be seen as the problem of finding the MAP (maximum a posteriori) in Bayesian modeling where Gaussian priors are used on columns of the factors $ U $ and $ V $, detailed discussion can be found in \cite{alquier2014bayesian,fithian2018flexible}.

Given an estimator $ \hat{M} $ as above,  the de-biased estimator \cite{chen2019inference} is defined as 
\begin{align}
M^{db} := {\rm Pr}_{{\rm rank-}r} \left[ \hat{M} - \mathcal{P}_\Omega ( \hat{M} - Y ) \right],
\end{align}
where ${\rm Pr}_{{\rm rank-}r} (B) = \arg\min_{A, {\rm rank}(A) \leq r} \| A- B\|_F  $ is the projection onto the set of rank-$r$ matrices.

\begin{remark}
	The estimation accuracy of the de-biased estimator, provided in Theorem 3 in \cite{chen2019inference} under some assumptions,  is 
	$
	\| M^{db} - M^* \|_F^2 \leq c \max(m,p)r \sigma^2/n
	$
	without any extra log-term and $c $ is universal numerical constant.
\end{remark}

\subsubsection{Confidence interval}
Let $ \hat{M} = \hat{U} \hat{\Sigma} \hat{V} ^{\top} $ be the singular values decomposition of $ \hat{M} $. Put 
\begin{align}
v_{ij} := \sigma^2 \left[ U^{db}_{i.} (U^{db\top}U^{db})^{-1}U^{db\top}_{i.} 
+
V^{db}_{j.} (V^{db\top}V^{db})^{-1}V^{db\top}_{j.} 
\right] /\kappa ,
\end{align}
where 
\begin{align*}
U^{db} =  \hat{U} (  \hat{\Sigma} + (\lambda/\kappa) I_r )^{1/2}
\text{ and }
V^{db} =  \hat{V} (  \hat{\Sigma} + (\lambda/\kappa) I_r )^{1/2}.
\end{align*}
Then, given a significance level $\alpha \in (0,1) $, the following interval
\begin{align*}
\left[ M^{db}_{ij} \pm \Phi^{-1} (1-\alpha/2) \sqrt{v_{ij}}  \right]
\end{align*}
is a nearly accurate two-sided $(1-\alpha)$ confidence interval of $M^*_{ij}$, where $  \Phi (\cdot) $ is the CDF of the standard normal distribution. This is given in Corollary 1 in \cite{chen2019inference}. This method is implemented in the \texttt{R} package \texttt{dbMC} \cite{dbMCpackage}.

\subsection{The Bayesian estimators}
The Bayesian estimator studied in  \cite{alquier2020concentration} is given by
$$
M^B := \int M \rho_\lambda ( M | Y ) dM
$$
where 
$$
\rho_\lambda ( M | Y )  \propto L(Y| M)^\lambda  \pi (M)
$$
is the posterior and $L(Y| M)^\lambda  $ is the likelihood raised to the power $\lambda$. Here $\lambda \in (0,1)$ is a tuning parameter and $ \pi (M)$ is the prior distribution.

\subsubsection{Priors}
A popular choice for the priors in Bayesian matrix completion is to assign conditional Gaussian priors to $U \in \mathbb{R}^{m\times K} $ and $V \in \mathbb{R}^{p\times K} $ such that
\begin{align*}
M = UV^\top = \sum_{k=1}^K U_{.k}V^\top_{.k},
\end{align*}
for a fixed integer $K \leq \min (m,p) $. More specifically,  for $k \in \{1,\ldots,K \} $, independently
\begin{equation}
\begin{aligned}
U_{.k}    & \sim \mathcal{N} (0,  \gamma_k I_m),
\\
V_{.k}    & \sim \mathcal{N} (0,  \gamma_k I_p),
\label{prior_full}
\\
\gamma_k^{-1}   & \sim \Gamma(a,b),
\end{aligned}
\end{equation}
where $ I_q $ is the identity matrix of dimension $q \times q $ and $a,b$ are some tuning parameters.  This type of prior is conjugate so that the conditional posteriors can be derived explicitly in closed form and allows to use the Gibbs sampler, see \cite{salakhutdinov2008bayesian} for details.  Some reviews and discussion on low-rank factorization priors can be found in \cite{alquier2013bayesian,alquier2014bayesian}.

\begin{remark}
	In the case that the rank $r$ is not known,  it is natural to take $K$ as large as possible,  e.g $K = \min(m,p) $ but this may be computationally prohibitive if $K$ is large.
\end{remark}

\begin{remark}
	The estimation error for this Bayesian estimator, under some assumptions, given in Corollary 4.2 in \cite{alquier2020concentration}, is 
	$
	\| M^{B} - M^* \|_F^2 \leq  \max(m,p)r \sigma^2/n
	$
	with an additional (multiplicative) log-term by $\log (n\max(m,p)) $. It is noted that the rate is also reached in \cite{mai2015} with an additional (multiplicative) log-term by $\log (\min(m,p)) $ under general sampling distribution however the authors considered some truncated priors.
\end{remark}

For a given rank-$r$, we propose to consider the following prior,  called fixed-rank-prior,
\begin{equation}
\begin{aligned}
U_{.k}    & \sim \mathcal{N} (0,  I_m),
\\
V_{.k}    & \sim \mathcal{N} (0, I_p),
\label{prior_fixed}
\end{aligned}
\end{equation}
for $k = 1, \ldots,r$. This prior is a simplified version of the above prior. We note that for $K > r$ the Gibbs sampler of the fixed-rank-prior will be faster than Gibbs sampler for the above prior. Interestingly, results from simulation for the Bayesian estimator with this prior are slightly better than the one based on the above prior at some point.

\begin{remark}
	We remark that the theoretical estimation error for the Bayesian estimator with the fixed-rank-prior given in \eqref{prior_fixed} remains unchanged following by Corollary 4.2 in \cite{alquier2020concentration}.
\end{remark}

\subsubsection{Credible intervals}
Using Bayesian approach, the credibility intervals for the matrix and their functions (e.g. entries) can be easily constructed using the Markov Chain Monte Carlo (MCMC) technique. Here, we focus on the equal-tailed  credible interval for an entry.  

More precisely,  the credible intervals are reported using the 89\% equal-tailed intervals that are recommended by \cite{kruschke2014doing,mcelreath2020statistical} for small posterior samples as in our situations with 500 posterior samples.  We noted that, according to \cite{salakhutdinov2008bayesian} as the data are too big to draw a reasonable size sample, the authors state that drawing only 500 observations from the Gibbs Sampler took 90 hours for the Netflix dataset.  Thus, we focus on the 89\% equal-tailed credible intervals for 500 posterior samples.  It is, however, noted that to obtain 95\% intervals,  an effective posterior sample size of at least 10.000 is recommended \cite{kruschke2014doing}, which is computationally costly to run on all of our simulations. A few examples with 10.000 posterior samples are examined in Figure \ref{fg_10000samples}.

\section{Simulation studies}
\label{sc_numerical}
\subsection{Experimental designs}
\label{sc_ex_setting}
In order to access the performance of different estimators, a series of experiments were conducted with simulated data. We fix $ m = 100 $ and alternate the other dimension by taking $ p = 100 $ and $ p = 1000 $. The rank $r$ is varied between $r = 2 $ and $r = 5 $.
\begin{itemize}
	\item Setting I: In the first setting, a rank-$r $ matrix
	$ M^*_{m\times p} $ is generated as the product of two rank-$r $ matrices, 
	$$ 
	M^* = U^*_{m\times r} V_{p\times r}^{*\top} ,
	$$
	where the entries of $U^* $ and $ V^* $ are i.i.d $ \mathcal{N} (0 , 1) $. With a missing rate $ \tau = 20\%, 50\% $ and $80\%$, the entries of the observed set are drawn uniformly at random. This sampled set is then corrupted by noise as in~\eqref{main model}, where the $ \mathcal{E}_i $ are i.i.d $ \mathcal{N} (0 , 1) $. 
	\item Setting II: The second series of simulations is similar to the first one, except that the matrix $M^*$ is no longer rank-$r$, but it can be well approximated by a rank-$r$ matrix:
	\[
	M^*=U^*_{m\times r} V_{p\times r}^{*\top} + \frac{1}{10}
	A_{m\times50} B_{p\times50}^\top
	\]
	where the entries of $A $ and $B$ are i.i.d $ \mathcal{N} (0 , 1) $.
	
	\item Setting III: This setting is similar to Setting I but here a heavy tail noise is used. More specifically, the noise $ \mathcal{E}_i $ are i.i.d Student distribution $ t_3 $ with 3 degrees of freedom. 
	
 	\item Setting IV: The set up of this setting is also similar to Setting I. However, we consider a more extreme case where the entries of $U^* $, $ V^* $  and the noise $ \mathcal{E}_i $ are all i.i.d drawn from the Student distribution with 3 degrees of freedom.
\end{itemize}

\begin{remark}
	We note that for the second series of simulations, with approximate low-rank matrices, the theory of the de-biased estimator can not be used whereas theoretical guarantees for Bayesian estimators are still valid, see \cite{alquier2020concentration}.
	The setting I follows exactly the minimax-optimal regime and thus it will allow to access the accuracy of the considered estimators. The last 2 settings (III and IV) are misspecification models set up where the theoretical guarantee is not available for all considered estimators.
\end{remark}

The behavior of an estimator (say $\widehat{M}$) is evaluated through the average squared error (ase) per entry
\[
{\rm ase} := \frac{1}{mp} \|\widehat{M} - M^*\|_F^2
\]
and the relative squared error (rse)
$$
{\rm rse} := \frac{ \|\widehat{M} - M^*\|_F^2}{  \| M^*\|_F^2 }.
$$
We also measure the error in predicting the missing entries by using
$$
{\rm Pred} := \frac{  \|   \mathcal{P}_{\bar{\Omega} } ( \widehat{M} - M^*)  \|_F^2}{mp- n},
$$
where $ \bar{\Omega} $ is the set of un-observed entries. For each setup, we generate 100 data sets (simulation replicates) and report the average and the standard deviation for a measure of error of each estimator over the replicates. 

We compare the de-biased estimator (denoted by `\textbf{d.b}'), the Bayesian estimator with the fixed-rank-prior \eqref{prior_fixed} (denoted by `\textbf{f.Bayes}') and the Bayesian estimator with the (flexible rank) prior \eqref{prior_full} (denoted by `\textbf{Bayes}'). As a by-product in calculating the de-biased estimator through the Alternating Least Squares estimator \eqref{estimator_als}, we also report the results for this estimator, denoted it by `\textbf{als}'. 

The `als' estimator is available from the  \texttt{R} package `\texttt{softImpute}' \cite{mazumder2010spectral} and is used with default options.  The `d.b' estimator is run with $ \lambda = 2.5\sigma\sqrt{mp} $ as in \cite{chen2019inference}. The `f.Bayes' and `Bayes' estimators are used with tuning parameter $\lambda = 1/(4\sigma^2) $ and parameters for the prior of 'Bayes' estimator are $K=10,a=1, b=1/100$.  The Gibbs samplers for these two Bayesian estimators are run with 500 steps and 100 burn-in steps.

\subsection{Results on estimation accuracy}
From the results in Tables \ref{tb_model1} and \ref{tb_model_approx}, it is clear that the de-biased estimator significantly outperforms its ancestry estimator being de-biased.  Whereas,  the de-biased estimator is just as good as the Bayesian methods in some cases.

More specifically,  in Table \ref{tb_model1}, the de-biased estimator behaves similar compared to Bayesian estimators in the case with high rates of observation (say $\tau = 20\%$ or $ 50\% $). With the case of highly missing rate $\tau = 80\% $,  the de-biased estimator returns highly unstable results, this may be because its ancestry estimator (here it is the als estimator) is unstable with few observations. However, when the dimension of the matrix increases, the differences between the de-biased estimator and the Bayesian estimators become smaller.  This is also recognized for the setting of approximate low-rank matrices as in Table \ref{tb_model_approx} and in Table \ref{tb_UQ}, \ref{tb_UQ_rankapprox}.

The `f.Bayes' method yields the best results quite often in terms of all considered errors (ase, Nase and Pred) in setting of exact low-rank matrices.  However,  it is noted that for the setting with the true underlying matrix being approximately low-rank, in Table \ref{tb_model_approx} and \ref{tb_UQ_rankapprox}, the `Bayes' approach is slightly better than the `f.Bayes' approach at some point. This can be explained as the `Bayes' approach employs a kind of approximate low-rank prior through the Gamma prior on the variance of the factor matrices and thus it is able to adapt to the approximate low-rankness.

Results in the cases of model misspecification with heavy tail noise are given Table \ref{tb_model_norstu} and \ref{tb_model_stustu}. Although Bayesian methods, especially `f.Bayes' method, yield better results compared with `als' or `db', all methods fail in the case of highly missing data, $\tau = 80\%$. This could be due to the fact that these considered methods are all designed for the case of Gaussian noise and thus they are not robust to other heavy tail noise, such as Student noise.

\subsection{Results on uncertainty quantification}
To examine the uncertainty quantification across the methods, we simulate a matrix as in Section \ref{sc_ex_setting} then we repeat the observation process 100 times.  More precisely, we obtain 100 data sets by replicating the observation of $ 20\%, 50\% $ and $80\%$ entries of the matrix $ M^* $ using a uniform sampling and then each sampled set is corrupted by noise as in \eqref{main model}, where the
$ \mathcal{E}_i $ are i.i.d $ \mathcal{N} (0 , 1) $.

Table \ref{tb_UQ} and \ref{tb_UQ_rankapprox} gather the empirical coverage rate of the confidence intervals and of the credible intervals of all methods over 100 independent experiments.  More precisely, we report the 95\% confidence intervals for the de-biased method. The credible interval is reported using the 89\% equal-tailed interval, see \cite{kruschke2014doing,mcelreath2020statistical}, for small posterior samples as in our situations with 500 posterior samples. We noted that to obtain 95\% intervals,  an effective posterior sample size of at least 10.000 is recommended \cite{kruschke2014doing}. A few examples from Setting I with 10000 posterior samples are given in Figure \ref{fg_10000samples}.

A noteworthy conclusion from the results in Table \ref{tb_UQ} and \ref{tb_UQ_rankapprox} is that the coverage rates of the 89\% credible intervals are significantly higher than those of the 95\% confidence intervals revealed by the de-biased method.  The credible intervals of the 'f.Bayes' approach show a slightly better coverage rate than those based on the 'Bayes' approach.  It is also noted that in the setting of approximate low-rankness,  Table \ref{tb_UQ_rankapprox}, where we do not have theoretical guarantee for the de-biased estimator,  the coverage rate of the confidence intervals is very low while the credible intervals still come with reliable coverage rates. These results further explain why Bayesian methods yield better results in accuracy as in Table \ref{tb_model1} and \ref{tb_model_approx}.

\begin{figure}[h]
	\centering
	\includegraphics[scale=.5]{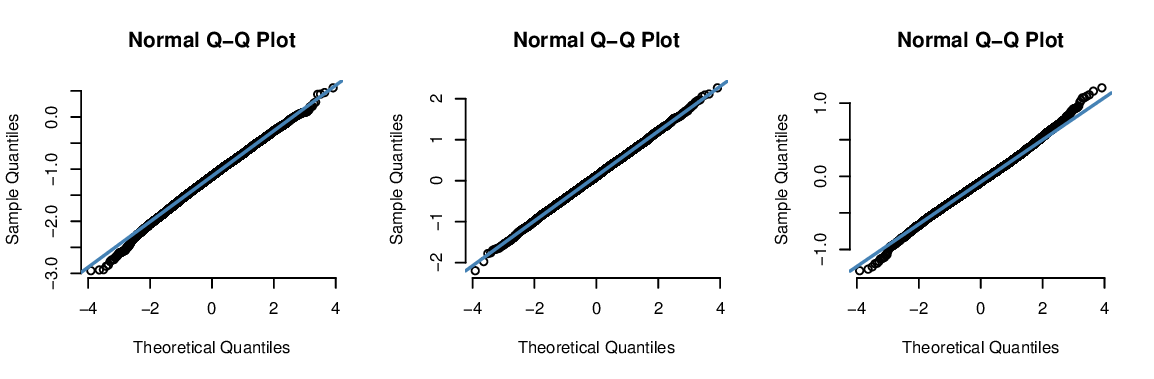}
	\includegraphics[scale=.5]{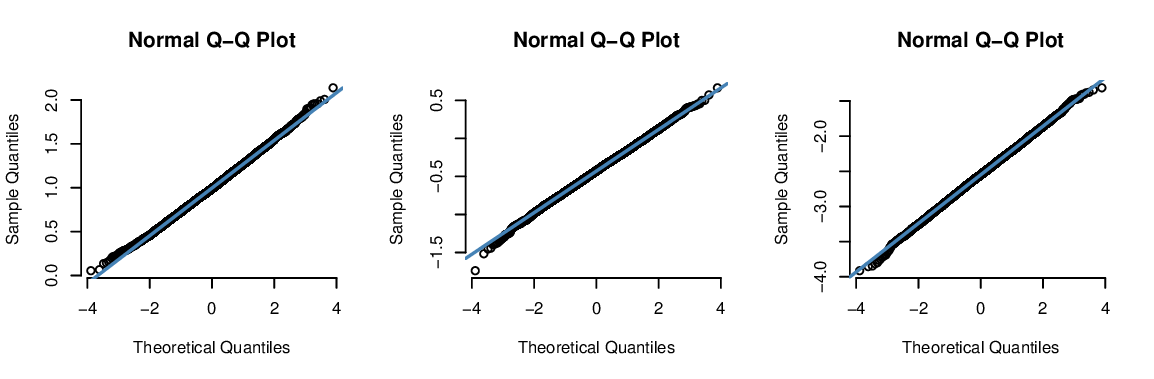}
	\caption{ Q-Q (quantile-quantile) plot to compare the 10000 posterior samples for some entries against the standard normal distribution.  Top row (from left to right, 3 figures) is the results from  Setting I with $r = 2,p = 100, \tau = 50\%$.  Bottom row (from left to right, 3 figures) is the results  from Setting I with $r = 2,p = 1000, \tau = 50\%$  }
	\label{fg_QQplot}
\end{figure}

\begin{figure}[h]
	\centering
	\includegraphics[scale=.5]{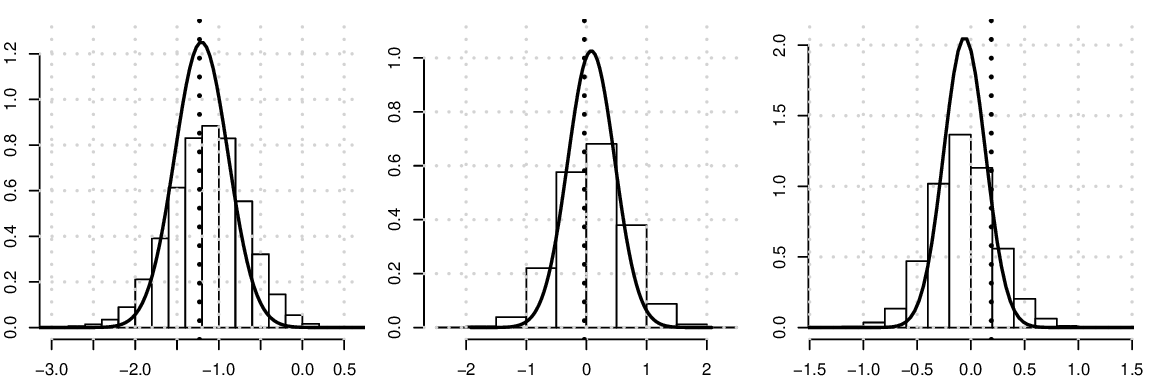}
	\includegraphics[scale=.5]{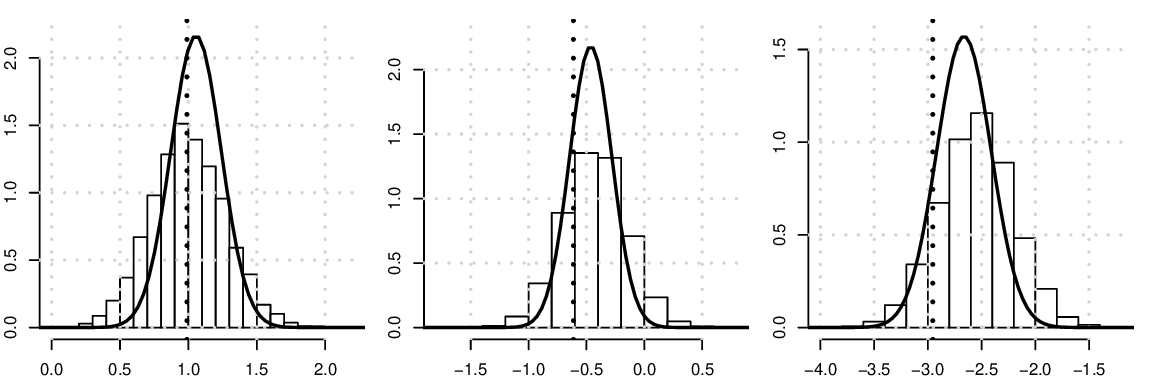}
	\caption{Plot to compare the limiting Gaussian distributions of the de-biased estimator and the histograms of the 10000 posterior samples for some entries.  The dotted line is the true value of the entries.  Top row (from left to right, 3 figures) is the results from  Setting I with $r = 2,p = 100, \tau = 50\%$.  Bottom row (from left to right, 3 figures) is the results  from Setting I with $r = 2,p = 1000, \tau = 50\%$  }
	\label{fg_10000samples}
\end{figure}

In Figure \ref{fg_10000samples}, we compare the limiting Gaussian distribution of the de-biased estimator and posterior samples for the 'f.Bayes' method against the true entries of interest.  It is shown that the limiting Gaussian distribution of the de-biased estimator yields a slightly sharper tail distribution compared to the distribution of the posterior samples.  In addition, Figure \ref{fg_QQplot} displays the Q-Q (quantile-quantile) plots of 10000 posterior samples of some entries vs. the standard Gaussian random variables. It shows that the posterior distributions of these entries reasonably well match the standard Gaussian distribution.

Results on empirical coverage rate of the confidence intervals and of the credible intervals for Setting III and IV with heavy tail noises are gathered in Table \ref{tb_UQ_norstu} and \ref{tb_UQ_stustu}. We can see that there is a slight reduction in the empirical coverage rate of all methods compared with those in the Gaussian noise setting in Table \ref{tb_UQ}. As in Setting I and II, the empirical coverage rates of confidence intervals decrease quickly as the missing rates $ \tau $ increase, while the empirical coverage rates of credible intervals remain stable.

\section{Discussion and Conclusion}
In this paper, we have provided extensive numerical comparisons between the de-biased estimator and the Bayesian estimators in the problem of low-rank matrix completion.  Results from numerical simulations draw a systematic picture of the behaviour of these estimators originally.   More specifically,  on the estimation accuracy, the de-biased estimator is comparable to the Bayesian estimators whereas the Bayesian estimators are much more stable and in some cases can outperform the de-biased estimator, especially in the small samples regime.  Moreover, the credible intervals reasonably cover the underlying entries quite well and slightly better than the confidence intervals in exact low-rank matrix completion.  However, in the case of approximate low-rankness,  the confidence intervals revealed by the de-biased estimators no longer work well. These results are interested for and can be served as a guideline for researchers as well as practitioners in many areas where one only has access to a few observations.

On the other hand, the results in this work suggest that the considered Bayesian estimators may actually reach the minimax-optimal rate of convergence without additional logarithmic factor. The extra log-terms could be due to the PAC-Bayesian bounds technique that used to prove the theoretical properties of the Bayesian estimator. Moreover, as shown in \cite{alquier2020concentration},  the same rate with log-term is proved for the concentration of the corresponding posterior and we conjecture that this rate could also be improved due to the coverage of credible intervals. These are important questions that remain open up to our knowledge.

Last but not least, it is also important to perform the comparisons with the Variational Bayesian (VB) method in \cite{lim2007variational} where its theoretical guarantees are given in \cite{alquier2020concentration}, because this method is very popular for matrix completion with large datasets.  This will be the objective of our future work. However, we would like to note that, in a preprint \cite{alquier2014bayesian}, the authors had performed some comparisons between the Bayesian approach and the VB method.  The message from their works is that we can expect that VB should be more or less as accurate as Bayes, maybe slightly less, but that the credibility intervals would be inaccurate (see e.g Figure 3 in \cite{alquier2014bayesian}).

\subsection*{Availability of data and codes}
The \texttt{R} codes and data used in the numerical experiments are available at:  \url{https://github.com/tienmt/UQMC} .

\subsection*{Acknowledgements}
	The author would like to thank the editor and the anonymous referee, who kindly reviewed the earlier version of the manuscript, for providing valuable suggestions and enlightening comments that help improve the current version of the paper. 
	TTM is supported by the Norwegian Research Council grant number 309960 through the Centre for Geophysical Forecasting at NTNU.
	The author would like to thank Pierre Alquier for kindly providing useful feedbacks on a first draft of this paper.

\begin{table}[!h]
	\tiny
	\caption{Simulation results for Setting I (exact low-rank). The mean and the standard deviation (in parentheses) of each error between the simulation replicates are presented.}
		\begin{tabular}{ l |cccc||cccc} 
			\hline\hline
			& \multicolumn{4}{ c | |}{$r = 2, p = 100,  \tau = 20\% $} 
			& \multicolumn{4}{ c  }{$r = 5, p = 100,  \tau = 20\% $} 
			\\
			Errors   & als & d.b & f.Bayes & Bayes 
			& als & d.b & f.Bayes & Bayes 
			\\ \hline
			ase	& 0.811 (.013) & 0.051 (.004) & 0.052 (.004) & 0.052 (.004)
			& 
			0.827 (.013) & 0.128 (.006) & 0.128 (.006) & 0.129 (.006)
			\\ 
			Nase  & 0.408 (.057) & 0.026 (.004) & 0.026 (.004) & 0.026 (.004)
			& 
			0.167 (.016)  & 0.026 (.003) & 0.026 (.003) & 0.026 (.003) 
			\\ 
			Pred  	& 0.055 (.005) & 0.055 (.005) & 0.055 (.004) & 0.055  (.004)
			& 
			0.144 (.008) & 0.144 (.008) & 0.145 (.009) & 0.145 (.010)
			\\
			\hline
			& \multicolumn{4}{ c ||}{  $r = 2, p = 100,  \tau = 50\% $  } 
			& \multicolumn{4}{ c  }{$r = 5, p = 100,  \tau = 50\% $} 
			\\  
			& als & d.b & f.Bayes & Bayes 
			& als & d.b & f.Bayes & Bayes 
			\\ \hline
			ase	&  0.548 (.011) & 0.088 (.006) & 0.089 (.007) & 0.089 (.007)
			& 
			0.634 (.014)  & 0.235 (.012) & 0.235 (.012)  & 0.236 (.012)
			\\
			rse	& 0.272 (.036) & 0.044 (.007) &  0.044  (.007) & 0.044 (.007)
			& 
			0.127 (.011) & 0.047 (.005) & 0.047 (.004) & 0.047 (.004)
			\\
			Pred  & 0.094 (.007) & 0.094 (.007) & 0.094 (.008)  & 0.095 (.008)
			& 
			0.268 (.015) & 0.268 (.015) & 0.267 (.016) & 0.269 (.016)
			\\
			\hline
			& \multicolumn{4}{ c| |}{  $r = 2, p = 100,  \tau = 80\% $  }
			& \multicolumn{4}{ c  }{ $r = 5, p = 100,  \tau = 80\% $}  
			\\  
			& als & d.b & f.Bayes & Bayes 
			& als & d.b & f.Bayes & Bayes 
			\\ \hline
			ase	& 0.476 (.168) & 0.320 (.178) & 0.288 (.028) & 0.298 (.031)
			& 
			3.522 (1.49)  & 3.474 (1.52) & 1.068 (.082) & 1.428 (.247)
			\\
			rse	& 0.244 (.085)  & 0.164 (.086) & 0.148 (.024)  & 0.153 (.029)
			& 
			0.713 (.311)  & 0.703 (.316) & 0.215 (.022) & 0.288 (.054)
			\\
			Pred  & 0.344 (.210) & 0.344 (.211) & 0.307 (.032)  & 0.317 (.035)
			&
			4.153 (1.87)  & 4.153 (1.87) & 1.204 (.098) & 1.613 (.285)
			\\ 
			\hline\hline
			& \multicolumn{4}{ c || }{$r = 2, p = 1000,  \tau = 20\% $} 
			& \multicolumn{4}{ c  }{$r = 5, p = 1000,  \tau = 20\% $} 
			\\
			& als & d.b & f.Bayes & Bayes 
			& als & d.b & f.Bayes & Bayes 
			\\ \hline
			 ase	& 0.806 (.004) & 0.028 (.001) & 0.028 (.001) & 0.028 (.001)
			& 
			0.814 (.004) & 0.070 (.001) &  0.070 (.001) & 0.071 (.001) 
			\\ 
			rse  & 0.407 (.041) & 0.014 (.001) & 0.014 (.001) & 0.014 (.001) 
			& 
			0.163 (.010) & 0.014 (.001) & 0.014 (.001) & 0.014 (.001) 
			\\ 
			Pred  	& 0.029 (.001) & 0.029 (.001)& 0.029 (.001) & 0.029 (.001) 
			& 
			0.076 (.002) & 0.076 (.002) & 0.076 (.002) & 0.076 (.002) 
			\\
			\hline
			& \multicolumn{4}{ c ||}{  $r = 2, p = 1000,  \tau = 50\% $  } 
			& \multicolumn{4}{ c  }{$r = 5, p = 1000,  \tau = 50\% $} 
			\\  
			& als & d.b & f.Bayes & Bayes 
			& als & d.b & f.Bayes & Bayes 
			\\ \hline
			 ase	& 0.523 (.004) & 0.046 (.001) & 0.046 (.001) & 0.046 (.001)
			& 
			0.564 (.004) & \textbf{0.119} (.002) & 0.120 (.002) & 0.120 (.002)
			\\ 
			rse	& 0.263 (.026) & 0.023 (.002) & 0.023 (.002) & 0.023 (.003) 
			& 
			0.112 (.007) & 0.024 (.002) & 0.024 (.001) & 0.024 (.001)
			\\
			Pred  & 0.048 (.002) & 0.048 (.002) & 0.048 (.002) & 0.048 (.002) 
			& 
			0.129 (.003) & 0.129 (.003) & 0.129 (.003) & 0.130 (.003) 
			\\
			\hline
			& \multicolumn{4}{ c| |}{  $r = 2, p = 1000,  \tau = 80\% $  }
			& \multicolumn{4}{ c  }{ $r = 5, p = 1000,  \tau = 80\% $}  
			\\  
			& als & d.b & f.Bayes & Bayes 
			& als & d.b & f.Bayes & Bayes 
			\\ \hline
		 ase & 0.344 (.324) & 0.167 (.331) & 0.135 (.005) & 0.136 (.005) 
			& 
			0.594 (.185) & 0.450 (.190) & \textbf{0.408} (.012) & 0.410 (.012) 
			\\
			rse	& 0.174 (.172) & 0.085 (.175) & 0.068 (.007) & 0.069 (.007) 
			& 
			0.120 (.040) & 0.091 (.041) & 0.082 (.012) & 0.082 (.011)
			\\
			Pred  & 0.180 (.404) & 0.180 (.406) & 0.141 (.006) & 0.142 (.006) 
			& 
			0.492 (.231) & 0.474 (.196) & \textbf{0.440} (.014) & 0.442 (.014)
			\\ \hline\hline
		\end{tabular}
	\label{tb_model1}
\end{table}

\begin{table}[!h]
	\tiny
	\caption{Simulation results for Setting II (approximate low-rank). The mean and the standard deviation (in parentheses) of each error between the simulation replicates are presented.}
		\begin{tabular}{ l |cccc ||cccc} 
			\hline\hline
			& \multicolumn{4}{ c || }{approximate rank-2, $p = 100,  \tau = 20\% $} 
			& \multicolumn{4}{ c  }{approximate rank-5, $p = 100,  \tau = 20\% $} 
			\\
			Errors   & als & d.b & f.Bayes & Bayes 
			& als & d.b & f.Bayes & Bayes 
			\\ \hline
			ase	& 0.914 (.013) & 0.533 (.017) & 0.533 (.017) & 0.532 (.017)	
			& 
			0.946 (.016) & 0.596 (.018) & 0.593 (.018) & 0.593 (.018)
			\\ 
			rse  & 0.370 (.042) & 0.216 (.026) & 0.215 (.026) & 0.215 (.026) 
			& 
			0.172 (.015) & 0.109 (.010) & 0.108 (.009) & 0.108 (.009)
			\\ 
			Pred  & 0.579 (.024) &  0.579 (.024) & 0.576 (.024) & 0.576 (.024) 
			& 
			0.724 (.034) & 0.724 (.034) &  0.716 (.033) & 0.716 (.033) 
			\\
			\hline
			& \multicolumn{4}{ c ||}{approximate rank-2, $ p = 100,  \tau = 50\% $  } 
			& \multicolumn{4}{ c  }{approximate rank-5, $p = 100,  \tau = 50\% $} 
			\\  
			& als & d.b & f.Bayes & Bayes 
			& als & d.b & f.Bayes & Bayes 
			\\ \hline
			ase	& 0.820 (.016) & 0.593 (.019) & 0.588 (.019) & 0.587 (.019)
			& 
			0.953 (.022) & 0.757 (.027) & 0.740 (.026) & 0.740 (.026)
			\\
			rse & 0.335 (.035) & 0.242 (.026) & 0.239 (.025) & 0.239 (.025) 
			& 
			0.173 (.015) & 0.138 (.012) & 0.134 (.012) & 0.135 (.012)
			\\
			Pred  & 0.642 (.026)  & 0.642 (.026) & 0.634 (.025) & 0.634 (.025)
			& 
			0.909 (.038) & 0.909 (.038) &  0.878 (.037) & 0.879 (.037)
			\\
			\hline
			& \multicolumn{4}{ c|| }{ approximate rank-2, $p = 100,  \tau = 80\% $  }
			& \multicolumn{4}{ c  }{approximate rank-5, $p = 100,  \tau = 80\% $}  
			\\  
			& als & d.b & f.Bayes & Bayes 
			& als & d.b & f.Bayes & Bayes 
			\\ \hline
			ase	& 1.258 (.760) & 1.189 (.777) & 0.839 (.041) &  0.842 (.042) 
			& 
			4.747 (1.04) & 4.603 (1.06) & 1.724 (.084) & 1.854 (.193)
			\\
			rse & 0.505 (.306) & 0.478 (.312) & 0.336 (.041) & 0.337 (.042)
			& 
			0.876 (.181) & 0.849 (.184) & 0.320 (.023) & 0.344 (.043)
			\\
			Pred  & 1.323 (.950) & 1.323 (.952) & 0.897 (.048) &  0.906 (.039)
			& 
			5.685 (1.30) & 5.305 (1.30) & 1.963 (.100) & 2.109 (.223)
			\\ 
			\hline\hline
			& \multicolumn{4}{ c || }{approximate rank-2, $p = 1000,  \tau = 20\% $} 
			& \multicolumn{4}{ c  }{approximate rank-5, $p = 1000,  \tau = 20\% $} 
			\\
			& als & d.b & f.Bayes & Bayes 
			& als & d.b & f.Bayes & Bayes 
			\\ \hline
			ase	& 0.909 (.005) & 0.521 (.010) & 0.521 (.010) & 0.500 (.010)		
			& 
			0.923 (.004) &  0.548 (.011) &  0.547 (.011) & 0.523 (.009)
			\\ 
			rse  & 0.368 (.030) & 0.211 (.017) & 0.211 (.017) & 0.203 (.017)	
			& 
			0.169 (.011) & 0.100 (.006) & 0.100 (.006) & 0.096 (.006)
			\\ 
			Pred  & 0.545 (.012) & 0.545 (.012) & 0.544 (.012) & 0.531 (.011)	
			& 
			0.610 (.014) & 0.610 (.014) & 0.608 (.014) & 0.591 (.011)
			\\
			\hline
			& \multicolumn{4}{ c ||}{ approximate rank-2, $p = 1000,  \tau = 50\% $  } 
			& \multicolumn{4}{ c  }{approximate rank-5, $p = 1000,  \tau = 50\% $} 
			\\  
			& als & d.b & f.Bayes & Bayes 
			& als & d.b & f.Bayes & Bayes 
			\\ \hline
			ase	& 0.786 (.007) & 0.546 (.012) & 0.545 (.012) & 0.544 (.011) 
			& 
			0.846 (.007) & 0.624 (.011) & 0.621 (.011) & 0.619 (.011) 
			\\
			rse & 0.315 (.025) & 0.219 (.018) & 0.219  (.018) & 0.218  (.018) 
			& 
			0.155 (.011) & 0.115 (.008) & 0.114 (.008) & 0.114 (.008)
			\\
			Pred  & 0.570 (.012) & 0.571 (.012) & 0.568  (.012) & 0.568  (.012) 
			& 
			0.694 (.013) & 0.694 (.013) & 0.687 (.013) & 0.687 (.013)
			\\
			\hline
			& \multicolumn{4}{ c|| }{ approximate rank-2, $p = 1000,  \tau = 80\% $  }
			& \multicolumn{4}{ c  }{ approximate rank-5, $p = 1000,  \tau = 80\% $}  
			\\  
			& als & d.b & f.Bayes & Bayes 
			& als & d.b & f.Bayes & Bayes 
			\\ \hline
			ase	& 0.770 (.013) & 0.681 (.015) & 0.666 (.014) & 0.666 (.015)
			& 
			1.149 (.079) & 1.077 (.081) & 0.992 (.021) & 0.992 (.021)
			\\
			rse & 0.316 (.027) & 0.280 (.024) & 0.274 (.023)  & 0.273 (.023)
			& 
			0.211 (.018) & 0.198 (.018) & 0.183 (.011) & 0.183 (.011)
			\\
			Pred  & 0.712 (.016) & 0.712 (.016) & 0.693 (.015) & 0.693 (.015)
			& 
			1.186 (.099) & 1.185 (.099) & 1.080 (.024) & 1.082 (.024)
			\\ 
			\hline\hline
		\end{tabular}
	\label{tb_model_approx}
\end{table}

\begin{table}[!h]
	\tiny
	\caption{Simulation results for Setting III (heavy tail noise). The mean and the standard deviation (in parentheses) of each error between the simulation replicates are presented. }
	\begin{tabular}{l|cccc||cccc} 
		\hline\hline
		& \multicolumn{4}{ c | |}{$r = 2, p = 100,  \tau = 20\% $} 
		& \multicolumn{4}{ c  }{$r = 5, p = 100,  \tau = 20\% $} 
		\\
		Errors   & als & d.b & f.Bayes & Bayes 
		& als & d.b & f.Bayes & Bayes 
		\\ \hline
		ase	& 2.568 (1.21) & 0.373 (1.25) & 0.236 (.415) & 0.511 (.313)
		& 2.696 (1.47) & 0.721 (1.51) & 0.552 (.675) & 0.689 (.341)
		\\ 
		rse & 1.292 (.672) & 0.193 (.663) & 0.120 (.212) & 0.255 (.163)
		& 0.543 (.303) & 0.145 (.306) & 0.111 (.138) & 0.139 (.072)
		\\ 
		Pred & 0.929 (4.92) & 0.932 (4.94) & 0.307 (.779) & 0.517 (.350)
		& 1.606 (6.19) & 1.612 (6.22) & 0.885 (2.02) & 0.812 (.454)
		\\
		\hline
		& \multicolumn{4}{ c ||}{  $r = 2, p = 100,  \tau = 50\% $  } 
		& \multicolumn{4}{ c  }{$r = 5, p = 100,  \tau = 50\% $} 
		\\  
		& als & d.b & f.Bayes & Bayes 
		& als & d.b & f.Bayes & Bayes 
		\\ 
		\hline
		ase	& 2.394 (2.29) & 1.147 (2.37) & 0.521 (.677) & 0.637 (.351)
		& 2.568 (1.83) & 1.525 (1.88) & 0.967 (1.11) & 1.050 (.912)
		\\
		rse & 1.249 (1.14) & 0.591 (1.18) & 0.271 (.333) & 0.335 (.183)
		& 0.518 (.367) & 0.307 (.374) & 0.194 (.218) & 0.211 (.179)
		\\
		Pred & 1.788 (4.26) & 1.793 (4.28) & 0.610 (.881) & 0.585 (.306)
		& 2.058 (2.80) & 2.064 (2.81) & 1.067 (.797) & 1.070 (.426)
		\\
		\hline
		& \multicolumn{4}{ c| |}{  $r = 2, p = 100,  \tau = 80\% $  }
		& \multicolumn{4}{ c  }{ $r = 5, p = 100,  \tau = 80\% $}  
		\\  
		& als & d.b & f.Bayes & Bayes 
		& als & d.b & f.Bayes & Bayes 
		\\ \hline
		ase	& 3.575 (1.91) & 3.273 (1.96) & 1.127 (.635) & 0.990 (.499)
		& 10.28 (1.97) & 10.22 (1.99) 
		& 2.603 (.599) & 2.487 (.574)
		\\
		rse & 1.837 (1.01) & 1.684 (1.02) & 0.576 (.336) & 0.505 (.262)
		& 2.063 (.436) & 2.051 (.439) 
		& 0.522 (.125) & 0.498 (.119)
		\\
		Pred  & 3.726 (2.31) & 3.732 (2.31) & 1.171 (.596) & 0.977 (.403)
		& 12.08 (2.38) & 12.09 (2.39) 
		& 2.830 (.565) & 2.682 (.516)
		\\ 
		\hline \hline
		& \multicolumn{4}{ c || }{$r = 2, p = 1000,  \tau = 20\% $} 
		& \multicolumn{4}{ c  }{$r = 5, p = 1000,  \tau = 20\% $} 
		\\
		& als & d.b & f.Bayes & Bayes 
		& als & d.b & f.Bayes & Bayes 
		\\ \hline
		ase	& 2.464 (.728) & 0.171 (.751) & 0.120 (.283) & 0.283 (.182)
		& 2.558 (.990) & 0.368 (1.04)
		& 0.290 (.604) & 0.389 (.188)
		\\ 
		rse & 1.255 (.415) & 0.089 (.404) & 0.062 (.151) & 0.144 (.096)
		& 0.511 (.194) & 0.073 (.202)
		& 0.059 (.129) & 0.078 (.039)
		\\ 
		Pred  & 0.414 (3.10) & 0.415 (3.12) & 0.177 (.786) & 0.293 (.210)
		& 0.825 (4.38) & 0.828 (4.40)
		& 0.482 (2.10) & 0.435 (.189)
		\\
		\hline
		& \multicolumn{4}{ c ||}{  $r = 2, p = 1000,  \tau = 50\% $  } 
		& \multicolumn{4}{ c  }{$r = 5, p = 1000,  \tau = 50\% $} 
		\\  
		& als & d.b & f.Bayes & Bayes 
		& als & d.b & f.Bayes & Bayes 
		\\ \hline
		ase	& 1.633 (.375) & 0.236 (.394) & 0.213 (.334) & 0.399 (.151)
		& 1.978 (1.13) & 0.685 (1.16)
		& 0.526 (.630) & 0.623 (.240)
		\\
		rse & 0.824 (.217) & 0.120 (.208) & 0.109 (.180) & 0.202 (.082)
		& 0.402 (.238) & 0.140 (.241)
		& 0.107 (.131) & 0.126 (.051)
		\\
		Pred & 0.284 (.622) & 0.285 (.625) & 0.250 (.513) & 0.391 (.142)
		& 0.918 (2.01) & 0.921 (2.02)
		& 0.631 (.976) & 0.656 (.216)
		\\
		\hline
		& \multicolumn{4}{ c| |}{  $r = 2, p = 1000,  \tau = 80\% $  }
		& \multicolumn{4}{ c  }{ $r = 5, p = 1000,  \tau = 80\% $}  
		\\  
		& als & d.b & f.Bayes & Bayes 
		& als & d.b & f.Bayes & Bayes 
		\\ \hline
		ase	& 1.250 (.700) & 0.756 (.715) & 0.590 (.644) & 0.631 (.348)
		& 2.161 (.570) & 1.778 (.579)
		& 1.200 (.280) & 1.221 (.170)
		\\
		rse & 0.624 (.334) & 0.376 (.337) & 0.291 (.292) & 0.313 (.153)
		& 0.434 (.123) & 0.357 (.123)
		& 0.241 (.060) & 0.245 (.038)
		\\
		Pred & 0.798 (.758) & 0.799 (.760) & 0.610 (.654) & 0.598 (.281)
		& 1.957 (.663) & 1.958 (.664)
		& 1.284 (.299) & 1.272 (.160)
		\\ \hline\hline
	\end{tabular}
	\label{tb_model_norstu}
\end{table}

\begin{table}[!h]
	\tiny
	\caption{Simulation results for Setting IV (extreme case). The mean and the standard deviation (in parentheses) of each error between the simulation replicates are presented.}
	\begin{tabular}{p{7mm}|cccc||cccc} 
		\hline\hline
		& \multicolumn{4}{ c | |}{$r = 2, p = 100,  \tau = 20\% $} 
		& \multicolumn{4}{ c  }{$r = 5, p = 100,  \tau = 20\% $} 
		\\
		Errors   & als & d.b & f.Bayes & Bayes 
		& als & d.b & f.Bayes & Bayes 
		\\ \hline
		ase	& 2.398 (.215) & 0.158 (.027) & 0.156 (.025) & 0.488 (.233) 
		& 2.812 (1.91) & 0.746 (1.89) & 0.559 (.927) & 0.791 (.522)
		\\ 
		rse  & 0.184 (.081) & 0.012 (.005) & 0.011 (.005) & 0.037 (.027) 
		& 0.071 (.035)  & 0.016 (.027) & 0.015 (.027) & 0.021 (.017) 
		\\ 
		Pred  	& 0.197 (.071) & 0.196 (.067) & 0.193 (.061) & 0.506  (.285)
		& 1.925 (8.89) & 1.894 (8.66) & 0.992 (2.69) & 0.950 (.484)
		\\
		\hline
		& \multicolumn{4}{ c ||}{  $r = 2, p = 100,  \tau = 50\% $  } 
		& \multicolumn{4}{ c  }{$r = 5, p = 100,  \tau = 50\% $} 
		\\  
		& als & d.b & f.Bayes & Bayes 
		& als & d.b & f.Bayes & Bayes 
		\\ 
		\hline
		ase	& 1.632 (.181) & 0.305 (.098) & 0.315 (.161) & 0.603 (.247)
		& 2.135 (.710)  & 0.978 (.696) & 0.930 (.566)  & 1.152 (.432)
		\\
		rse & 0.120 (.049) & 0.022 (.010) &  0.021  (.010) & 0.044 (.025)
		& 0.056 (.021) & 0.025 (.016) & 0.023 (.011) & 0.030 (.013)
		\\
		Pred  & 0.371 (.184) & 0.368 (.179) & 0.392 (.309)  & 0.615 (.293)
		& 1.301 (1.10) & 1.293 (1.10) & 1.237 (1.00) & 1.382 (.659)
		\\
		\hline
		& \multicolumn{4}{ c| |}{  $r = 2, p = 100,  \tau = 80\% $  }
		& \multicolumn{4}{ c  }{ $r = 5, p = 100,  \tau = 80\% $}  
		\\  
		& als & d.b & f.Bayes & Bayes 
		& als & d.b & f.Bayes & Bayes 
		\\ \hline
		ase	& 6.777 (7.64) & 6.420 (7.75) & 3.721 (5.19) & 1.700 (1.39)
		& 19.54 (6.17)  & 19.45 (6.23) & 6.559 (2.96) & 5.713 (2.63)
		\\
		rse	& 0.485 (.597)  & 0.458 (.605) & 0.211 (.236)  & 0.112 (.058)
		& 0.597 (.189)  & 0.595 (.191) & 0.169 (.052) & 0.148 (.041)
		\\
		Pred  & 7.718 (9.53) & 7.724 (9.55) & 4.399 (6.39)  & 1.845 (1.69)
		& 23.70 (7.70)  & 23.68 (7.71) & 7.774 (3.63) & 6.699 (3.23)
		\\ 
		\hline \hline
		& \multicolumn{4}{ c || }{$r = 2, p = 1000,  \tau = 20\% $} 
		& \multicolumn{4}{ c  }{$r = 5, p = 1000,  \tau = 20\% $} 
		\\
		& als & d.b & f.Bayes & Bayes 
		& als & d.b & f.Bayes & Bayes 
		\\ \hline
		ase	& 2.441 (.401) & 0.115 (.276) & 0.109 (.242) & 0.336 (.429)
		& 2.478 (.201) & 0.229 (.081) & 0.220 (.031) & 0.423 (.220) 
		\\ 
		rse  & 0.159 (.063) & 0.008 (.024) & 0.008 (.022) & 0.022 (.036) 
		& 0.061 (.014) & 0.005 (.001) & 0.005 (.001) & 0.010 (.005) 
		\\ 
		Pred  	& 0.149 (.360) & 0.146 (.352) & 0.133 (.289) & 0.377 (.492) 
		& 0.315 (.397) & 0.309 (.373) & 0.268 (.075) & 0.516 (.238) 
		\\
		\hline
		& \multicolumn{4}{ c ||}{  $r = 2, p = 1000,  \tau = 50\% $  } 
		& \multicolumn{4}{ c  }{$r = 5, p = 1000,  \tau = 50\% $} 
		\\  
		& als & d.b & f.Bayes & Bayes 
		& als & d.b & f.Bayes & Bayes 
		\\ \hline
		ase	& 1.680 (.815) & 0.244 (.842) & 0.353 (1.98) & 0.464 (.254)
		& 1.799 (.337) & 0.474 (.306) & 0.419 (.121) & 0.664 (.210)
		\\
		rse & 0.110 (.036) & 0.013 (.023) & 0.016 (.054) & 0.030 (.015) 
		& 0.045 (.011) & 0.011 (.003) & 0.010 (.004) & 0.016 (.006)
		\\
		Pred  & 0.322 (1.38) & 0.322 (1.39) & 0.542 (3.66) & 0.468 (.211) 
		& 0.613 (.600) & 0.605 (.584) & 0.501 (.186) & 0.758 (.236) 
		\\
		\hline
		& \multicolumn{4}{ c| |}{  $r = 2, p = 1000,  \tau = 80\% $  }
		& \multicolumn{4}{ c  }{ $r = 5, p = 1000,  \tau = 80\% $}  
		\\  
		& als & d.b & f.Bayes & Bayes 
		& als & d.b & f.Bayes & Bayes 
		\\ \hline
		ase	& 6.350 (15.1) & 5.880 (15.3) & 6.414 (18.1) & 0.849 (.346) 
		& 6.426 (10.1) & 6.028 (10.2) & 2.287 (2.96) & 2.246 (2.20)
		\\
		rse	& 0.351 (.794) & 0.321 (.802) & 0.328 (.895) & 0.051 (.015) 
		& 0.127 (.133) & 0.117 (.135) & 0.056 (.085) & 0.050 (.017)
		\\
		Pred  & 7.188 (18.9) & 7.198 (18.9) & 7.887 (22.6) & 0.876 (.403)
		& 7.294 (12.6) & 0.474 (.196) & 2.646 (3.68) & 2.528 (2.74)
		\\ \hline\hline
	\end{tabular}
	\label{tb_model_stustu}
\end{table}

\begin{table}
	\small
	\caption{Simulation results for Setting I on empirical coverage rate of the confidence intervals and of the credible intervals of the entries (standard deviation is given in parentheses).} 
	\centering
		\begin{tabular}{  c |  c  c c  }
			\hline\hline
			& \multicolumn{3}{ c }{empirical coverage rate (\%) } 
			\\ 
			missing rate
			&  CI db & CI  f.Bayes  & CI Bayes 
			\\ 
			\hline
			& \multicolumn{3}{ c}{  $r = 2, p = 100$  }
			\\ 
			\hline
			$ \tau = 20\% $ & 99.9 (0.1)  & 99.9 (0.1) & 99.3 (0.3) 
			\\
			$ \tau = 50\% $ & 94.1 (1.1) & 99.9 (0.1) & 99.2 (0.3) 
			\\
			$ \tau = 80\% $ & 58.9 (4.8) & 98.6 (0.4) & 98.5 (0.5)
			\\ 
			\hline
			& \multicolumn{3}{ c}{  $r = 2, p = 1000$  }
			\\ \hline   %%%%%%%%%%%%%%%%%%
			$ \tau = 20\% $ &  99.9 (0.1) & 99.9 (0.1) & 99.4 (0.1)
			\\
			$ \tau = 50\% $ &  94.5 (0.4) & 99.9 (.01) &  99.3 (0.1)
			\\
			$ \tau = 80\% $ & 64.4 (0.9) & 99.6 (0.1) & 99.0 (0.2)
			\\ \hline
			& \multicolumn{3}{ c}{  $r = 5, p = 100 $  }
			\\ 
			\hline  %%%%%%%%%%%%%%%%
			$ \tau = 20\% $ & 99.9 (0.1) & 99.9 (0.1) & 99.2 (0.2)
			\\
			$ \tau = 50\% $ & 93.0 (1.3) & 99.8 (0.1) &  99.1 (0.2)
			\\
			$ \tau = 80\% $ & 38.5 (4.5) & 96.4 (0.6) & 90.9 (2.8)
			\\ 
			\hline
			& \multicolumn{3}{ c}{  $r = 5, p = 1000$  }
			\\ \hline   %%%%%%%%%%%%%%%%%%
			$ \tau = 20\% $ &  99.9 (0.1) & 99.9 (0.1) & 99.4 (0.1)
			\\
			$ \tau = 50\% $ &  94.1 (0.3) & 99.9 (0.1) &  99.2 (0.1)
			\\
			$ \tau = 80\% $ & 60.4 (0.8) & 99.2 (0.1) & 98.7 (0.1)
			\\ \hline \hline
		\end{tabular}
	\label{tb_UQ}
\end{table}

\begin{table}[!ht]
	\small
	\caption{Simulation results for Setting II on empirical coverage rate of the confidence intervals and of the credible intervals of the entries (standard deviation is given in parentheses).} 
	\centering
		\begin{tabular}{  c |  c  c c  }
			\hline\hline
			& \multicolumn{3}{ c  }{empirical coverage rate (\%) } 
			\\ 
			missing rate
			&  CI db & CI f.Bayes  & CI  Bayes 
			\\ 
			\hline
			& \multicolumn{3}{ c}{ approximate rank-2, $p = 100$  }
			\\ \hline
			$ \tau = 20\% $ & 70.4 (0.3) & 69.3 (0.3) & 55.9 (0.4)
			\\
			$ \tau = 50\% $ & 49.3 (0.4) & 78.0 (0.4) & 65.2 (0.5) 
			\\
			$ \tau = 80\% $ & 33.0 (2.6) & 89.5 (0.6) & 81.3 (0.7)
			\\ \hline
			& \multicolumn{3}{ c}{ approximate rank-2, $p = 1000$  }
			\\ \hline   %%%%%%%%%%%%%%%%%%
			$ \tau = 20\% $ & 57.2 (0.1) & 56.0 (0.1) & 51.1 (1.8)
			\\
			$ \tau = 50\% $ & 37.5 (1.3) & 63.0 (0.2) & 51.2 (1.1)
			\\
			$ \tau = 80\% $ & 27.6 (0.8) & 77.6 (0.3) & 67.3 (0.4)
			\\ \hline
			& \multicolumn{3}{ c}{ approximate rank-5, $p = 100 $  }
			\\ \hline  %%%%%%%%%%%%%%%%
			$ \tau = 20\% $ & 90.1 (0.3) & 88.9 (0.3) & 76.9 (0.4)
			\\
			$ \tau = 50\% $ & 66.9 (0.5) & 93.2 (0.3) & 83.8 (0.4)
			\\
			$ \tau = 80\% $ & 30.9 (3.1) & 94.2 (0.6) & 90.2 (2.8)
			\\ 
			\hline
			& \multicolumn{3}{ c}{  approximate rank-5, $p = 1000$  }
			\\ \hline   %%%%%%%%%%%%%%%%%%
			$ \tau = 20\% $ & 78.9 (0.1) & 77.4 (0.1) & 69.1 (1.1)
			\\
			$ \tau = 50\% $ & 56.3 (0.2) & 85.0 (0.1) & 73.4 (0.9)
			\\
			$ \tau = 80\% $ & 38.0 (0.7) & 92.6 (0.2) & 86.2 (0.3)
			\\ \hline \hline
		\end{tabular}
	\label{tb_UQ_rankapprox}
\end{table}

\begin{table}
	\small
	\caption{Simulation results for Setting III on empirical coverage rate of the confidence intervals and of the credible intervals of the entries (standard deviation is given in parentheses).} 
	\centering
	\begin{tabular}{  l |  c  c c  }
		\hline\hline
		& \multicolumn{3}{ c }{empirical coverage rate (\%) } 
		\\ 
		missing rate
		&  CI db & CI  f.Bayes  & CI Bayes 
		\\ \hline
		& \multicolumn{3}{ c}{  $r = 2, p = 100$  }
		\\ \hline
		$ \tau = 20\% $ & 95.7 (5.2) & 96.4 (4.8) & 97.9 (0.4) 
		\\
		$ \tau = 50\% $ & 71.4 (7.6) & 94.9 (4.6) & 98.2 (0.3) 
		\\
		$ \tau = 80\% $ & 30.6 (5.8) & 94.9 (2.0) & 97.1 (0.8) 
		\\ \hline
		& \multicolumn{3}{ c}{  $r = 2, p = 1000$  }
		\\ \hline   %%%%%%%%%%%%%%%%%%
		$ \tau = 20\% $ & 96.7 (4.5) & 96.7 (3.4) & 97.2 (0.4) 
		\\
		$ \tau = 50\% $ & 75.6 (3.1) & 95.9 (4.5) & 97.7 (0.3) 
		\\
		$ \tau = 80\% $ & 41.7 (4.0) & 95.5 (2.0) & 98.5 (0.2) 
		\\ \hline
		& \multicolumn{3}{ c}{  $r = 5, p = 100 $  }
		\\ \hline  %%%%%%%%%%%%%%%%
		$ \tau = 20\% $ & 95.9 (3.1) & 96.4 (2.5) & 95.2 (0.8) 
		\\
		$ \tau = 50\% $ & 70.5 (3.2) & 96.5 (1.6) & 95.7 (0.7) 
		\\
		$ \tau = 80\% $ & 24.6 (2.3) & 95.3 (0.8) & 94.0 (1.6) 
		\\ \hline
		& \multicolumn{3}{ c}{  $r = 5, p = 1000$  }
		\\ \hline   %%%%%%%%%%%%%%%%%%
		$ \tau = 20\% $ & 96.9 (1.7) & 96.8 (2.5) & 93.9 (0.6) 
		\\
		$ \tau = 50\% $ & 74.9 (1.8) & 96.6 (1.6) & 94.6 (0.4) 
		\\
		$ \tau = 80\% $ & 39.6 (2.1) & 96.1 (0.6) & 95.9 (0.5) 
		\\ \hline \hline
	\end{tabular}
	\label{tb_UQ_norstu}
\end{table}

\begin{table}
	\small
	\caption{Simulation results for Setting IV on empirical coverage rate of the confidence intervals and of the credible intervals of the entries (standard deviation is given in parentheses).} 
	\centering
	\begin{tabular}{  c |  c  c c  ccc }
		\hline\hline
		& \multicolumn{3}{ c  }{empirical coverage rate (\%) } 
		\\ 
		missing rate
		&  CI db & CI  f.Bayes  & CI Bayes 
		\\ \hline
		& \multicolumn{3}{ c}{  $r = 2, p = 100$  }
		\\ \hline
		$ \tau = 20\% $ & 97.1 (1.1) & 96.4 (4.7) & 97.9 (0.5) 
		\\
		$ \tau = 50\% $ & 75.4 (4.9) & 95.6 (6.3) & 98.1 (0.3) 
		\\
		$ \tau = 80\% $ & 33.4 (7.1) & 94.7 (2.8) & 97.4 (0.7) 
		\\ \hline
		& \multicolumn{3}{ c}{  $r = 2, p = 1000$  }
		\\ \hline   %%%%%%%%%%%%%%%%%%
		$ \tau = 20\% $ & 97.0 (0.6) & 97.2 (0.4) & 97.2 (0.4) 
		\\
		$ \tau = 50\% $ & 76.0 (4.1) & 97.2 (0.4) & 97.9 (0.3) 
		\\
		$ \tau = 80\% $ & 39.1 (6.0) & 93.8 (5.9) & 98.6 (0.2) 
		\\ \hline
		& \multicolumn{3}{ c}{  $r = 5, p = 100 $  }
		\\ \hline  %%%%%%%%%%%%%%%%
		$ \tau = 20\% $ & 97.0 (0.9) & 96.9 (2.2) & 95.7 (0.7) 
		\\
		$ \tau = 50\% $ & 72.4 (2.7) & 96.7 (0.9) & 96.0 (0.6) 
		\\
		$ \tau = 80\% $ & 22.9 (3.3) & 92.9 (1.4) & 94.5 (1.0) 
		\\ \hline
		& \multicolumn{3}{ c}{  $r = 5, p = 1000$  }
		\\ \hline   %%%%%%%%%%%%%%%%%%
		$ \tau = 20\% $ & 97.1 (0.3) & 97.2 (0.2) & 94.6 (0.5) 
		\\
		$ \tau = 50\% $ & 75.3 (0.7) & 96.9 (1.2) & 94.8 (0.4) 
		\\
		$ \tau = 80\% $ & 37.3 (2.6) & 96.2 (0.5) & 96.3 (0.3) 
		\\ \hline \hline
	\end{tabular}
	\label{tb_UQ_stustu}
\end{table}

%\begin{acknowledgements}
%\end{acknowledgements}


\section*{Conflict of interest}
 The authors declare that they have no conflict of interest.

\bibliographystyle{apalike}      % basic style, author-year citations
%\bibliography{refs_MC}   % name your BibTeX data base

%\begin{thebibliography}{}
% etc
%\end{thebibliography}

\appendix

\end{document}